%% file: main_final.tex
\documentclass[conference]{IEEEtran}
\IEEEoverridecommandlockouts

\usepackage{amsmath,amssymb,amsfonts}
\usepackage{graphicx}
\usepackage{url}
\usepackage{algorithm}
\usepackage{algpseudocode}
\usepackage{algorithmicx}
\usepackage{booktabs}
\usepackage{multicol}
\usepackage{multirow}
\usepackage{colortbl}
\usepackage{siunitx}
\usepackage{cite}
\usepackage{placeins}
\usepackage[svgnames]{xcolor}

\algrenewcommand\algorithmicrequire{\textbf{Require:}}
\algrenewcommand\algorithmicensure{\textbf{Ensure:}}

\begin{document}

\title{JEDI: JEPA-to-Edge Distillation for Efficient\\
Cropland Segmentation from Satellite Imagery}

\author{%
  Kishor Kumar Bhaumik \quad Nicolas Roque dos Santos \quad Jia Chen \quad Evangelos E. Papalexakis \\
  University of California, Riverside \\
  \texttt{\normalsize \{kbhau001,nicolasr,jiac,epapalex\}@ucr.edu}
}

\maketitle

\begin{abstract}

Large vision models provide meaningful representations for remote-sensing
segmentation but are often too expensive for deployment at the satellite
or field edge. Knowledge distillation can reduce this cost. However, existing feature-level methods have two key limitations. First, they typically assume the teacher and student share a similar architecture. Second, they usually stop aligning features once task training begins, even though that alignment is what made the student's features useful in the first place. We therefore introduce \textbf{JEDI} (JEPA-to-Edge
Distillation), a two-stage framework that transfers an I-JEPA
(Image-based Joint-Embedding Predictive Architecture) representation
from a large isotropic Vision Transformer (ViT) teacher to a student
model with substantially fewer parameters while preserving the
teacher's prediction capacity. JEDI first aligns the student's terminal
representation with the teacher's token space through a simple
cross-architecture projection and spatial alignment. The aligned
student is then jointly optimized using supervised segmentation,
temperature-scaled response distillation and a persistent feature
alignment objective that remains active throughout task adaptation. On the CalCROP21 dataset,
JEDI-B0 reaches 68.0 mean Intersection-over-Union (mIoU) with only
4.04M parameters. This is a 16.0 point gain over the standalone
student and comes within 2.0 points of the 70.0 mIoU achieved by the
639M-parameter teacher. We adopt SegFormer as the student family for its promising accuracy-to-parameter ratio on segmentation. We evaluate three variants of SegFormer --- B0, B1, and B2 with 4.04M, 14.33 M, and 28M parameters respectively. Across all three, JEDI consistently outperforms response-, structure-, channel- and relational-distillation baselines under the same teacher--student setting. Overall, these results show that persistent representation alignment matters most under aggressive compression, substantially cutting model size and compute while preserving performance. Our code is available at
\textcolor{VioletRed}{\url{https://github.com/Kishor-Bhaumik/JEDI}}.

\end{abstract}

\section{Introduction}

Accurate and timely delineation of cultivated land is important for
agricultural monitoring and public-sector decision making. The USDA Cropland Data Layer (CDL), for example, provided annual crop maps at \SI{30}{\meter} resolution through year 2023. From year 2024 onward, it has provided \SI{10}{\meter} resolution maps. However, these products remain retrospective. Consequently, they are not designed for applications that require information during the growing season~\cite{ghosh2021calcrop21}. Recent work has shown that
freely available satellite imagery, particularly Sentinel-2, can support
more timely crop mapping at substantially finer spatial resolution
\cite{defourny2019near}. A systematic review of remote-sensing crop mapping further shows that deep learning has become an important component of large-scale crop mapping~\cite{alami2023crop}. However, early-season mapping, multi-temporal reasoning and reliable use of multi-source imagery remain open challenges.These trends
motivate models that can exploit modern visual representations while
remaining practical for repeated inference over large geographic areas.
\begin{figure}[t]
    \centering
    \includegraphics[width=\columnwidth]{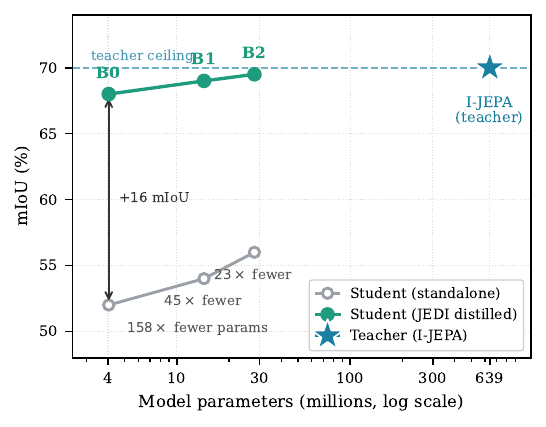}
    \caption{JEDI distills a $639$M-parameter I-JEPA teacher into
    compact SegFormer students. The smallest student (B0, $158\times$
    fewer parameters than teacher) gains $+16$ mIoU over its non-distilled baseline,
    reaching within $2.0$ mIoU points of the teacher.}
    \label{fig:teaser}
\end{figure}
The deployment setting makes this problem substantially different from
conventional semantic segmentation. Satellite and field-edge platforms
operate under tighter constraints on memory, power and computation than
datacenter GPUs, while the volume of imagery can make transmitting raw
observations itself expensive. The $\Phi$-Sat-1 mission~\cite{giuffrida2021varphi} demonstrated the
feasibility of running deep neural networks directly on an earth
observation satellite, illustrating the potential of moving part of the
analysis closer to the sensor. At the same
time, remote-sensing segmentation has its own challenges, including
large spatial variation, substantial intra-class variability and
foreground--background imbalance that are not fully captured by methods
developed for natural images~\cite{zheng2020foreground}. Consequently, a
model that is highly accurate on a conventional GPU is not necessarily
an appropriate model for resource-constrained earth observation.

One approach is to use increasingly capable vision foundation models.
Self-supervised learning has substantially improved the quality of
visual representations available for downstream dense prediction.
DINO~\cite{caron2021emerging}, DINOv2~\cite{oquab2023dinov2} and
MAE~\cite{he2022masked} learn transferable representations from unlabeled
images, while satellite-specific methods adapt these objectives to the
spectral and temporal structure of Earth observation data. SatMAE, for
example, introduces temporal embeddings and spectral positional
encodings specifically for multi-spectral and temporal satellite
imagery and reports improvements on downstream land-cover and
segmentation tasks~\cite{cong2022satmae}. More recently, geospatial
foundation models such as Prithvi have demonstrated that large-scale
self-supervised pretraining on Harmonized Landsat--Sentinel-2 imagery
can transfer to diverse Earth observation tasks, including crop-related
applications~\cite{jakubik2023foundation}. These developments
suggest that large pretrained representations can provide useful semantic
structure for remote-sensing applications; however, their size can make
direct deployment impractical.

A natural response is to compress a large model into a smaller one using
knowledge distillation. Response-based distillation transfers softened
predictions~\cite{hinton2015distilling}, while feature-based methods
transfer information from intermediate representations
\cite{srivastava2015training,tian2019contrastive}. For semantic segmentation, the
structured nature of dense predictions has motivated methods that
transfer spatial relations, channel distributions and cross-image
relationships~\cite{liu2019structured,shu2021channel,yang2022cross}. Other work
has focused on preserving intra-class feature variation
\cite{wang2020intra} and recent studies show that even direct raw-feature
distillation can be highly competitive when the feature magnitudes and
distillation weights are carefully controlled~\cite{liu2024rethinking}.
These results establish knowledge distillation as an effective mechanism
for recovering part of a large model's accuracy with a substantially
smaller student.

However, most feature-based distillation methods assume that the teacher
and student expose representations that can be matched directly or
stage-by-stage. This assumption becomes problematic when the two
networks differ not only in capacity but also in architecture.
Notably, recent work has explicitly considered heterogeneous
CNN--Transformer distillation, demonstrating that useful information can
be transferred between models with different representation
characteristics \cite{zhu2023good}. Nevertheless, the problem becomes
more severe when the teacher is a large isotropic ViT (e.g., ViT-H/16, as
used by I-JEPA) producing a single token sequence and the student is a
compact hierarchical Transformer (e.g., SegFormer) producing a
multi-scale feature pyramid. In this setting, the teacher
and student representations differ in embedding dimension, spatial
resolution and topology. Simply applying a conventional feature loss
therefore requires either discarding spatial information or introducing a
substantial transformation whose effect on the student representation
is difficult to control.

We address this problem with \textbf{JEDI} (JEPA-to-Edge Distillation),
a two-stage cross-architecture distillation framework that transfers
representation knowledge from a large I-JEPA teacher to a compact
SegFormer student. In the first stage, we explicitly align the
student's terminal representation with the teacher's token space using
a lightweight learned projection and spatial resampling, allowing the
student to acquire the teacher's representation without segmentation
supervision. In the second stage, the aligned student is jointly
optimized using supervised segmentation, temperature-scaled response
distillation and a persistent feature-alignment objective. Unlike a
conventional pretrain-then-fine-tune strategy, the feature objective is
not removed after initialization; it remains active while the student
adapts to the segmentation task. This design is motivated by the
observation that aggressive compression can cause task supervision to
drive the student away from useful teacher representations.

We further reduce inference cost by selecting, for each tile, a single observation from its multi-temporal Sentinel-2 stack. Specifically, this selection is guided by a criterion based on the Normalized Difference Vegetation Index (NDVI), a standard remote-sensing measure of vegetation greenness computed from the near-infrared and red spectral bands. This selection is applied
identically during training and at inference: the segmentation network
always operates on one NDVI-selected composite per tile rather than the
full temporal sequence, so the same lightweight acquisition-selection
mechanism that reduces training cost also reduces the cost of deployed
inference. This does not attempt to replace temporal modeling; instead,
it avoids evaluating the segmentation network repeatedly over the
complete temporal stack. The resulting system therefore targets a
practical accuracy--compute trade-off rather than maximum segmentation
accuracy under unconstrained computation. Our main contributions are:
\begin{itemize}
    \item We propose \textbf{JEDI}, a cross-architecture distillation
    framework for transferring representations between a large
    isotropic I-JEPA ViT and a compact hierarchical SegFormer whose
    feature spaces differ in both dimensionality and spatial topology.

    \item We introduce a two-stage training strategy that first performs
    representation alignment and then jointly optimizes segmentation,
    response distillation and persistent feature alignment, retaining
    representational supervision throughout task adaptation.

    \item We demonstrate on the CalCROP21 dataset that the resulting compact students
    substantially outperform identically configured non-distilled
    students, with the $4.04$M-parameter JEDI-B0 reaching $68.0$ mIoU,
    within $2.0$ points of the approximately $639$M-parameter teacher.

        \item We characterize the resulting accuracy--compute trade-off
    across multiple student capacities using parameter count,
    multiply--accumulate operations (GMACs) and GPU inference latency.
\end{itemize}
Fig.~\ref{fig:teaser} summarizes this accuracy--compute
trade-off across the teacher and the JEDI student variants.

\section{Related Work}
\label{sec:related}

\subsection{Cropland Mapping and Remote-Sensing Segmentation}

Remote-sensing crop mapping has evolved from traditional
hand-engineered spectral and temporal features toward supervised machine
learning and deep neural networks. A recent systematic review covering
hundreds of crop-mapping studies identifies deep learning as a major
direction while highlighting persistent challenges in early crop
mapping, crop rotation, multi-source imagery and reliable ground-truth
construction~\cite{alami2023crop}. These challenges are
particularly relevant for satellite-based segmentation because crop
appearance changes with phenology, acquisition conditions and geographic
context.

At the architectural level, semantic segmentation has progressed from
fully convolutional networks~\cite{long2015fully} and encoder--decoder
architectures such as U-Net~\cite{ronneberger2015u} to architectures
that explicitly model multi-scale context, including DeepLab
\cite{chen2018encoder}. Transformer-based models subsequently
introduced global attention into visual representation learning
\cite{dosovitskiy2020image}, while Mask2Former~\cite{cheng2022masked}
provided a unified formulation for several dense-prediction tasks.
Remote-sensing imagery, however, introduces additional difficulties
beyond those encountered in natural-image segmentation. FarSeg, for
example, explicitly models foreground--scene relations and addresses
foreground--background imbalance and large intra-class variation in
geospatial object segmentation~\cite{zheng2020foreground}. These observations
underscore the importance of representations that preserve semantic
structure while remaining robust to spatial and appearance variation.

For crop mapping specifically, the availability of Sentinel-2 has enabled
multi-temporal and multi-spectral approaches that exploit spectral
signatures and phenological changes. CalCROP21 provides a representative
benchmark by combining Sentinel-2 observations over California's Central
Valley with crop labels derived from and refined relative to the USDA
Cropland Data Layer~\cite{ghosh2021calcrop21}. Our work uses this benchmark
but focuses on a different question: rather than maximizing segmentation
accuracy with an unconstrained model, we investigate how much of a large
teacher's representation can be transferred to a substantially smaller
student without significantly sacrificing the peformance of segmentation.

\subsection{Satellite Representation Learning and Foundation Models}

The success of self-supervised learning in natural images has motivated
the development of representation learning specifically adapted to earth observation. DINO~\cite{caron2021emerging} and DINOv2~\cite{oquab2023dinov2}
learn transferable visual features through self-supervised
objectives, while MAE~\cite{he2022masked} learns representations by
reconstructing masked image content. While such approaches achieve promising representation learning performance on RGB imagery, they are not suitable for satellite imagery due to its spectral and temporal dimensions.


SatMAE~\cite{cong2022satmae} adapts masked autoencoding to this setting by incorporating
temporal embeddings and spectral positional encodings for satellite
imagery. Its results show that explicitly modeling the
structure of multi-spectral and temporal observations can improve
transfer to downstream remote-sensing tasks, including semantic
segmentation. Geospatial foundation models extend this direction further.
Prithvi, for example, is pretrained on more than one terabyte of
Harmonized Landsat--Sentinel-2 imagery and is designed for transfer
across multiple Earth observation tasks, including flood mapping,
wildfire-scar segmentation and crop-related applications
\cite{jakubik2023foundation}. These models demonstrate the value of large
self-supervised representations for Earth observation, but their
increasing capacity also strengthens the need for efficient transfer
mechanisms when the final model must operate under constrained compute
or memory budgets.

I-JEPA~\cite{assran2023self} provides a self-supervised formulation by
predicting representations of target image regions from context, rather
than reconstructing pixels as in existing masked-image or contrastive
methods. This latent predictive objective avoids requiring the
representation to reproduce low-level pixel details and instead
encourages prediction in an abstract embedding space. JEDI uses I-JEPA
as the teacher and investigates how such a large latent representation
can be transferred into a compact hierarchical segmentation model. This
distinguishes our work from methods that directly fine-tune a large
foundation model: the objective is not to deploy the foundation model
itself but to use its learned representation to train a substantially
smaller model such as smaller variants of SegFormer~\cite{xie2021segformer}.

\subsection{Efficient Segmentation and Edge Deployment}

The computational cost of modern segmentation models has motivated
architectures that explicitly balance accuracy and efficiency.
SegFormer~\cite{xie2021segformer} combines a hierarchical Mix Transformer
encoder with a lightweight all-MLP decoder and provides multiple model
sizes, including the compact MiT-B0 configuration used in our study.
Such architectures reduce the cost of the deployed model by design,
rather than relying solely on post-training compression.

Efficiency is especially important for Earth observation because
inference can be performed close to the sensor or on field-edge
platforms with limited power and memory. ESA's $\Phi$-Sat-1 demonstrated
in-orbit deep-learning inference using an Intel Myriad~2 vision
processing unit, illustrating the feasibility of onboard processing for
Earth observation~\cite{giuffrida2021varphi}. Subsequent work has
investigated aggressive compression and hardware-aware neural
architecture search for in-orbit inference~\cite{furano2020towards,
del2025optimizing}. These efforts primarily optimize or compress a given
architecture for a target hardware platform. In contrast, JEDI focuses
on a different source of efficiency: transferring the representation of
a large teacher into a compact student whose architecture and
representation topology are substantially different.

This distinction is important for our setting. A large pretrained teacher
can provide useful semantic structure but may be too expensive to
execute repeatedly over large satellite image collections. Rather than
requiring the deployed model to retain the teacher architecture, we
treat the teacher as a training-time source of representation knowledge
and discard it after distillation.

\subsection{Knowledge Distillation for Semantic Segmentation}

Knowledge distillation~\cite{hinton2015distilling} provides a general
mechanism for transferring information from a 
complex
teacher model to
a compact student model. Response-based distillation uses the teacher's
softened output distribution as additional supervision, while
feature-based approaches transfer information from intermediate
representations~\cite{srivastava2015training,tian2019contrastive}. The latter is
particularly attractive for semantic segmentation because dense
prediction contains spatial and semantic structure that may not be
fully represented by the final logits.

Several methods therefore design segmentation-specific forms of
distillation. Structured Knowledge Distillation (SKD) models dense
prediction as a structured labeling problem and transfers pairwise
relationships between pixels~\cite{liu2019structured}. Channel-wise Knowledge
Distillation (CWD) treats channel activations as distributions and
matches their statistics between teacher and student
\cite{shu2021channel}. IFVD instead focuses on intra-class feature variation,
using class-wise feature centers to transfer the distributional
structure learned by the teacher~\cite{wang2020intra}. CIRKD extends
relational distillation across images by transferring global
pixel-to-pixel and pixel-to-region relationships
\cite{yang2022cross}. These approaches illustrate a progression from
matching predictions to transferring increasingly structured properties
of the teacher representation.

More recent work has revisited the assumptions behind feature
distillation. Liu et al.~\cite{liu2024rethinking} show that direct
distillation of raw features can remain highly effective and analyze
the role of feature magnitude and angular information in determining
distillation behavior. Other work has explicitly considered
heterogeneous architectures. Zhu et al.~\cite{zhu2023good} investigate
knowledge exchange between CNN and Transformer models through
heterogeneous feature distillation and selective prediction
distillation. Such work is particularly relevant to JEDI because it
demonstrates that teacher--student pairs need not share the same
architectural family.

Nevertheless, existing feature-distillation methods generally assume
that the teacher and student representations can be compared after
relatively simple correspondence or transformation. In our setting, the
I-JEPA teacher is an isotropic ViT producing a single token sequence,
whereas the SegFormer student is hierarchical and produces a
multi-resolution feature pyramid. The two representations therefore
differ simultaneously in dimensionality, spatial resolution and
topology. 
We propose JEDI to address this mismatch explicitly through a learned
cross-architecture alignment module. In addition, conventional
distillation pipelines often use feature transfer as an initialization
or auxiliary training signal whose importance decreases once task
supervision is introduced. JEDI instead keeps the feature-alignment
objective active throughout task adaptation. Our experiments isolate
these two choices and evaluate their contribution under substantial
model compression.

\begin{table}[t]
\centering
\caption{Representation shapes used throughout the framework. Student
stage widths are given for the MiT-B0\,/\,B1\,/\,B2 variants.}
\label{tab:dims}
\renewcommand{\arraystretch}{1.2}
\begin{tabular}{@{}ll@{}}
\toprule
Quantity & Value \\
\midrule
Input $x$                       & $3 \times 448 \times 448$ \\
Classes $K$                     & $2$ \\
Teacher tokens $z_T$            & $784 \times 1280$ \\
Teacher grid                    & $28 \times 28$ \\
Student pyramid resolutions     & $112^2,\ 56^2,\ 28^2,\ 14^2$ \\
Terminal-stage width $C_4$      & $256\,/\,512\,/\,512$ \\
Student logits $\hat{y}_S$      & $K \times 112 \times 112$ \\
\bottomrule
\end{tabular}
\end{table}

\section{Methodology}
\label{sec:method}

\subsection{Problem Formulation and Notation}

\begin{figure*}[t]
    \centering
    \includegraphics[width=\textwidth,keepaspectratio]{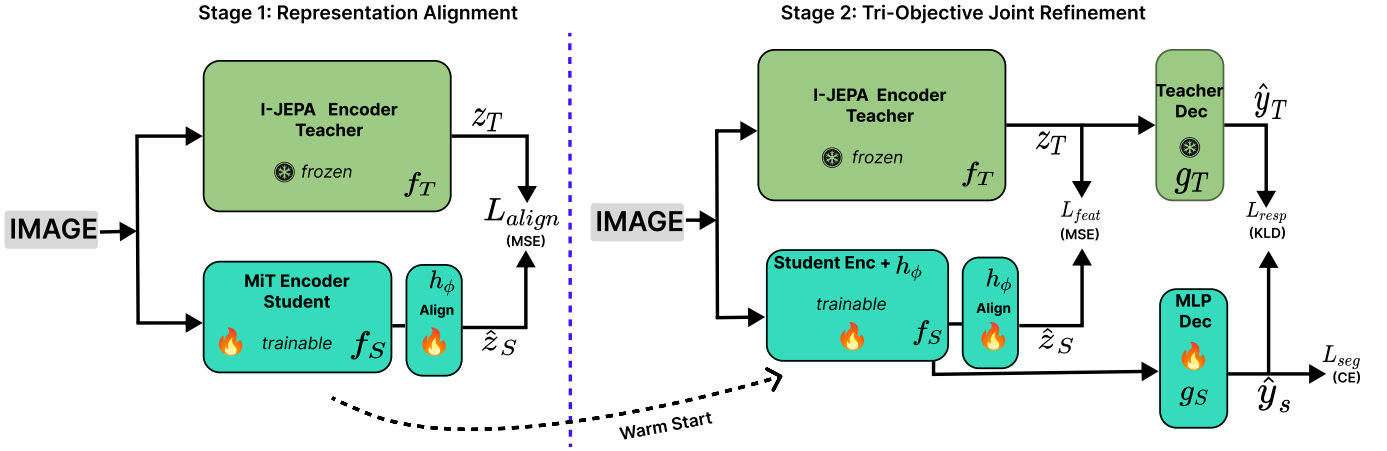}
    \caption{The JEDI framework. \textbf{Stage~1 (left):} the trainable
    MiT student encoder is aligned to the frozen I-JEPA teacher, with
    the alignment module $h_\phi$ lifting the student's terminal-stage
    features into the teacher's representation space so that a single
    feature-reconstruction objective $\mathcal{L}_{\mathrm{align}}$ can
    be applied. No task labels are used. \textbf{Stage~2 (right):} the
    student is warm-started from Stage~1 and adapted to segmentation
    under three concurrent objectives---supervised segmentation
    ($\mathcal{L}_{\mathrm{seg}}$), response distillation from the
    teacher's predictions ($\mathcal{L}_{\mathrm{resp}}$) and a
    persistent feature-alignment term ($\mathcal{L}_{\mathrm{feat}}$)
    that keeps the student's representation from drifting away from the
    teacher's as it specializes. The teacher, decoder and $h_\phi$ are
    all discarded at inference; only $f_S$ and $g_S$ are deployed.}
    \label{fig:workflow}
\end{figure*}

Let $\mathcal{X} \in \mathbb{R}^{T \times C \times H \times W}$ denote a
multi-temporal Sentinel-2 observation stack over a fixed geographic tile,
comprising $T$ acquisitions across $C$ spectral bands at spatial
resolution $H \times W$. Let $y$ denote the corresponding pixel-wise
cropland annotation. Fig.~\ref{fig:workflow} illustrates the overall JEDI pipeline which spans both training stages.

Our goal is to train a compact student network $\mathcal{S}$ on a
single selected observation by transferring knowledge from a high-capacity
teacher $\mathcal{T}$. The temporal stack is used only during acquisition
selection; the deployed segmentation model receives the resulting
three-channel observation.

We write $f_T$ and $g_T$ for the teacher encoder and decoder, $f_S$ and
$g_S$ for their student counterparts and $h_\phi$ for the
cross-architecture alignment module introduced in
Section~\ref{subsec:align}. Latent token representations are denoted
$z$, intermediate feature maps $F$ and class logits $\hat{y}$.
Trainable parameters are $\theta_S$ (student encoder), $\psi$ (student
decode head) and $\phi$ (alignment module); teacher parameters remain
frozen throughout. Table~\ref{tab:dims} collects the tensor shapes
referenced below.

\subsection{NDVI-Guided Composite Selection}
\label{subsec:ndvi}

Processing the full temporal stack through the segmentation network
would multiply the neural-network inference cost by the number of
acquisitions. We therefore select a single acquisition exhibiting strong
vegetative expression. For each timestep $t$ we compute the Normalized
Difference Vegetation Index from the near-infrared and red
spectral bands,

\begin{equation}
\mathrm{NDVI}_t(i,j) =
\frac{\mathcal{X}_{t,\mathrm{NIR}}(i,j) - \mathcal{X}_{t,\mathrm{RED}}(i,j)}
     {\mathcal{X}_{t,\mathrm{NIR}}(i,j) + \mathcal{X}_{t,\mathrm{RED}}(i,j)},
\end{equation}

guarding against division by zero and select

\begin{equation}
t^{\star} = \arg\max_{t \in \{1,\dots,T\}}
\frac{1}{HW}\sum_{i,j} \mathrm{NDVI}_t(i,j).
\end{equation}

The visible bands of acquisition $t^{\star}$ are cropped and standardized
to form the network input $x$. This reduces the number of neural-network
evaluations from $T$ to one while retaining a lightweight preprocessing
step for acquisition selection. The criterion also tends to favor
low-cloud acquisitions, since cloud cover generally depresses scene-level
NDVI, while the resulting three-channel input remains compatible with
the RGB-pretrained teacher.

\subsection{Teacher and Student Models}
\label{subsec:models}

The teacher encoder $f_T$ is an I-JEPA ViT-H/16 backbone
($\approx\!639$M parameters) pretrained by latent-space block
prediction~\cite{assran2023self} and subsequently fine-tuned for
cropland segmentation on the target domain. Fine-tuning is
performed on the same CalCROP21 training split used for the student, by attaching the
convolutional decoder $g_T$ and optimizing both $f_T$ and $g_T$ with a
binary cross-entropy loss for $50$ epochs. Being isotropic, it emits a
flat, single-scale token sequence $z_T = f_T(x)$, which the fine-tuned
lightweight convolutional decoder $g_T$ upsamples to a full-resolution
prediction $\hat{y}_T$. Both $f_T$ and $g_T$ are frozen in all subsequent
stages and contribute no gradients.

The student adopts the SegFormer design~\cite{xie2021segformer}: a
hierarchical Mix Transformer (MiT) encoder $f_S$ producing a four-stage
feature pyramid $\{F^{(1)},\dots,F^{(4)}\}$ of progressively coarser,
deeper maps, paired with a lightweight all-MLP decode head $g_S$ that
fuses the four stages into logits $\hat{y}_S$ at quarter resolution. We
evaluate three capacity points: MiT-B0, B1 and B2, spanning $4.04$M
to $28$M parameters against the $639$M-parameter teacher.

We tap the terminal stage $F^{(4)}$ for distillation. This choice is
deliberate: the teacher emits a single representation at one semantic
abstraction level and $F^{(4)}$ is the deepest, most abstract student
stage, whose receptive field and degree of abstraction most closely
correspond. Matching a shallower, higher-resolution stage
would instead force the student to reproduce semantically abstract
targets from features that have not yet accumulated sufficient context,
a mismatch the alignment module cannot fully resolve.

\subsection{Cross-Architecture Alignment Module}
\label{subsec:align}

Direct feature distillation between $F^{(4)}$ and $z_T$ is
ill-posed, since the two representations disagree in width, in spatial
resolution and in layout, contrasting a channel-first feature map with a
flat token sequence. The alignment module $h_\phi$ resolves all three
mismatches through a composition of three operations, producing a
target-matched student representation

\begin{equation}
\hat{z}_S = h_\phi(F^{(4)}).
\label{eq:align_module}
\end{equation}

It first \emph{tokenizes} the terminal feature map by flattening its
spatial dimensions and transposing it into the teacher's sequence
convention. It then \emph{projects} each token through a single linear
map into the teacher's embedding width. Finally, it bilinearly resamples
the resulting feature map to the teacher's spatial resolution before
flattening it into a sequence, yielding exact position-by-position
correspondence with the teacher tokens.

Two design choices are important for this alignment. First, the
projection is kept linear and shallow. A more expressive projector could
absorb part of the alignment problem itself, allowing the encoder to
satisfy the loss without substantially improving its representations.
Restricting $h_\phi$ to an affine map places the representational burden
on $f_S$, which is the component ultimately retained at inference.
Because $h_\phi$ is discarded after training, it contributes no inference
cost. Second, we resample the \emph{student} upward rather than pooling
the \emph{teacher} downward. Downsampling the teacher would discard
spatial information from the supervisory target, whereas upsampling
preserves the teacher's full spatial grid.

\subsection{Stage 1: Representation Alignment}
\label{subsec:stage1}

The first stage trains the student encoder to reproduce the teacher's
latent geometry with no task supervision whatsoever. Given the frozen
teacher targets $z_T = f_T(x)$, we minimize

\begin{equation}
\mathcal{L}_{\mathrm{align}}(\theta_S, \phi)
= \frac{1}{N_T D_T}
\left\| h_\phi\big(f_S(x)\big) - z_T \right\|_F^2
\label{eq:align}
\end{equation}

over $\theta_S$ and $\phi$ jointly, where $N_T D_T$ normalizes by the
teacher representation size. The decode head is excluded
from this stage and receives no gradient, since the objective concerns
representation rather than prediction. Admitting task-specific parameters
here would let the student begin specializing before it has acquired the
representation that specialization is meant to build upon.

We use squared error rather than cosine similarity so that both the
direction and magnitude of the teacher representation contribute to the
alignment objective. This provides a direct reconstruction target in the
teacher's embedding space without introducing an additional normalization
operation. Upon convergence, $\theta_S$ and $\phi$ are checkpointed to
initialize the second stage.

\subsection{Stage 2: Tri-Objective Joint Refinement}
\label{subsec:stage2}

The second stage adapts the aligned student to the segmentation task
while continuing to constrain its representation. The student encoder
and alignment module are warm-started from Stage~1; the decode head
retains its ImageNet initialization. All three components are trained.

A single encoder pass per network supplies every quantity required. The
student pyramid $\{F^{(s)}\}$ yields both the projected tokens
$\hat{z}_S = h_\phi(F^{(4)})$ and the logits
$\hat{y}_S = g_S(\{F^{(s)}\})$; likewise the teacher tokens $z_T$ are
reused by $g_T$ to produce $\hat{y}_T$ without a second traversal of the
teacher backbone. This shared-computation formulation avoids redundant
encoder evaluations when constructing the three loss terms.

We optimize a weighted composition of three terms.

\emph{Task supervision.} Since $\hat{y}_S$ is predicted at quarter
resolution, we downsample the ground-truth mask to $\tilde{y}$ using
nearest-neighbor interpolation to preserve discrete class labels. The
segmentation loss is the standard cross-entropy over the prediction grid
$\Omega$:

\begin{equation}
\mathcal{L}_{\mathrm{seg}} =
-\frac{1}{|\Omega|}\sum_{p \in \Omega}
\log
\frac{\exp\big(\hat{y}_S^{(\tilde{y}_p)}(p)\big)}
     {\sum_{k=1}^{K}\exp\big(\hat{y}_S^{(k)}(p)\big)}.
\end{equation}

\emph{Response distillation.} Teacher logits are bilinearly resampled
to the student lattice, $\hat{y}_T^{\downarrow}$ and matched in
softened probability space with temperature $\tau$:

\begin{equation}
\mathcal{L}_{\mathrm{resp}} = \tau^2 \cdot
\mathrm{KL}\!\left(
\sigma\big(\hat{y}_T^{\downarrow}/\tau\big)
\ \big\|\
\sigma\big(\hat{y}_S/\tau\big)
\right),
\end{equation}

with $\sigma$ the channel-wise softmax and the $\tau^2$ factor restoring
gradient scale~\cite{hinton2015distilling}. This term transfers the
teacher's inter-class uncertainty, which can be informative at field
boundaries and over spectrally ambiguous regions where hard labels may
be less reliable.

\emph{Persistent feature alignment.} The Stage~1 objective is retained
unchanged:

\begin{equation}
\mathcal{L}_{\mathrm{feat}} =
\frac{1}{N_T D_T}\left\| \hat{z}_S - z_T \right\|_F^2 .
\end{equation}

Finally, the composite objective is

\begin{equation}
\mathcal{L} =
\alpha\,\mathcal{L}_{\mathrm{seg}} +
\beta\,\mathcal{L}_{\mathrm{resp}} +
\gamma\,\mathcal{L}_{\mathrm{feat}} .
\label{eq:total}
\end{equation}

Retaining $\mathcal{L}_{\mathrm{feat}}$ during task adaptation is the
central design choice of JEDI. Stage~1 establishes an initialization in
the teacher's representation space, but task supervision can subsequently
shift the student toward features that are sufficient for the segmentation
labels without preserving the teacher representation. Keeping
$\gamma>0$ makes representational similarity an explicit constraint
throughout adaptation rather than only at initialization. The
$\gamma=0$ ablation in Section~\ref{sec:experiments} isolates the effect
of removing this persistent constraint.

At inference, the teacher, the alignment module and the projection are
all discarded. The deployed model comprises only $f_S$ and $g_S$.
Algorithm~\ref{alg:jedi} summarizes the complete two-stage
JEDI training procedure.

\begin{algorithm}[t]
\caption{JEDI Training Procedure}
\label{alg:jedi}
\begin{algorithmic}[1]
\Require Stacks $\{\mathcal{X}_i\}$, labels $\{y_i\}$; frozen teacher $(f_T, g_T)$
\Ensure Deployable student $(f_S, g_S)$
\State $x_i \leftarrow$ \textsc{NdviComposite}$(\mathcal{X}_i)$ for all $i$
\State \textbf{// Stage 1: representation alignment}
\For{epoch $= 1$ to $E_1$}
  \For{minibatch $x$}
    \State $z_T \leftarrow f_T(x)$ \Comment{no grad}
    \State $\hat{z}_S \leftarrow h_\phi(f_S(x))$
    \State update $\theta_S, \phi$ by $\nabla \mathcal{L}_{\mathrm{align}}$ \Comment{Eq.~\eqref{eq:align}}
  \EndFor
\EndFor
\State \textbf{// Stage 2: tri-objective joint refinement}
\For{epoch $= 1$ to $E_2$}
  \For{minibatch $(x, y)$}
    \State $z_T \leftarrow f_T(x)$;\ \ $\hat{y}_T \leftarrow g_T(z_T)$ \Comment{single pass}
    \State $\{F^{(s)}\} \leftarrow f_S(x)$ \Comment{single pass}
    \State $\hat{z}_S \leftarrow h_\phi(F^{(4)})$;\ \ $\hat{y}_S \leftarrow g_S(\{F^{(s)}\})$
    \State update $\theta_S, \psi, \phi$ by $\nabla \mathcal{L}$ \Comment{Eq.~\eqref{eq:total}}
  \EndFor
\EndFor
\State \Return $(f_S, g_S)$ \Comment{discard $f_T, g_T, h_\phi$}
\end{algorithmic}
\end{algorithm}

\section{Experiments}
\label{sec:experiments}

\subsection{Experimental Setup}

\textbf{Dataset.} We evaluate on CalCROP21~\cite{ghosh2021calcrop21}, which
pairs multi-temporal Sentinel-2 acquisitions over California's Central
Valley with CDL-derived labels refined by a spatio-temporal attention
model. We cast the task as binary cropland segmentation (cultivated
versus non-cultivated) and adopt the train/test partition of
\cite{gurav2023can}; all model-selection decisions are fixed a priori
from the configuration described below. Each multi-temporal tile is
reduced to a single three-channel composite by the NDVI-guided selection
of Section~\ref{subsec:ndvi} and center-cropped to $448 \times 448$,
matching the fixed input resolution expected by both the
teacher and student networks (Table~\ref{tab:dims}).

\textbf{Implementation.} All models are implemented in PyTorch and
trained on a single NVIDIA RTX~A6000 (48\,GB). We use the AdamW
optimizer with a cosine-annealed learning rate (initial
$10^{-4}$, weight decay $0.05$) and gradient clipping at unit norm.
Stage~1 runs for $E_1=30$ epochs and Stage~2 for $E_2=20$ epochs at batch
size $8$. We set The distillation temperature $\tau = 2$ and the
ViT-H/16 backbone fine-tuned for cropland segmentation and frozen
Stage~2 loss weights $(\alpha, \beta, \gamma) = (1.0, 0.5, 0.5)$ . The teacher is an I-JEPA thereafter.

\textbf{Metrics.} Segmentation quality is reported as mean
Intersection-over-Union (mIoU) and macro F1. Efficiency is characterized
by parameter count, multiply--accumulate operations (GMACs) at
$448\times448$ input and inference latency measured on a single A6000
GPU. GMACs count the multiply-accumulate operations (each
counted once, regardless of the two floating-point operations it
comprises) performed during a single forward pass, and thus serve as a
hardware-independent proxy for a model's computational cost at inference
time. All segmentation results are reported as mean $\pm$ standard
deviation over five independent seeds.

\subsection{Baselines}
\label{subsec:baselines}

We compare JEDI against four prior distillation methods spanning the
main families of the field, with the non-distilled student as a lower
reference and the fine-tuned teacher as an upper reference. All methods
use the same frozen teacher and student architectures under an otherwise
identical training protocol, differing in their distillation objectives.
Because several feature-level baselines were originally developed for
architectures with spatially compatible feature maps, we route their
feature-level objectives through the same alignment module described in
Section~\ref{subsec:align}. This provides a common representation space
for the teacher--student comparison.\\
\textbf{KD}~\cite{hinton2015distilling} acts as a baseline by aligning
temperature-scaled output distributions between the teacher and student
networks.\\
\textbf{SKD}~\cite{liu2019structured} transfers pairwise pixel similarities and
a holistic term, exploiting the structured nature of dense prediction.\\
\textbf{CWD}~\cite{shu2021channel} normalizes each channel into a
distribution and matches teacher and student channels by KL divergence.\\
\textbf{CIRKD}~\cite{yang2022cross} provides a relational distillation
baseline that extends distillation beyond single images by matching
cross-image pixel relationships between teacher and student.\\\\
The baseline comparison therefore evaluates whether the proposed
cross-architecture alignment and persistent feature objective provide
additional benefit beyond established response-, structure-, channel-,
and relational-distillation objectives.
\input{table_main}

\subsection{Main Results}

Table~\ref{tab:main} presents the main results on CalCROP21 across
three student capacities. The results show a consistent improvement
from standalone training to knowledge distillation and, further, to
JEDI. For MiT-B0, the standalone model obtains $52.0\pm0.6$ mIoU and
$66.0\pm0.5$ F1. Conventional response-based KD improves these results
to $60.5\pm0.5$ mIoU and $73.0\pm0.4$ F1, while the segmentation-specific
distillation methods SKD, CWD and CIRKD progressively increase mIoU
to $62.8\pm0.5$, $64.7\pm0.5$ and $65.9\pm0.5$, respectively. JEDI
achieves $68.0\pm0.4$ mIoU and $80.0\pm0.3$ F1, improving over the
standalone baseline by $16.0$ mIoU points and $14.0$ F1 points.
Furthermore, it outperforms the strongest prior
distillation baseline, CIRKD, by $2.1$ mIoU points and $2.3$ F1 points.

The same ordering is observed for the larger students. For MiT-B1,
standalone training obtains $54.0\pm0.6$ mIoU and $67.5\pm0.5$ F1,
whereas CIRKD reaches $67.2\pm0.4$ mIoU and $78.8\pm0.4$ F1. JEDI
further improves these results to $69.0\pm0.4$ mIoU and $81.0\pm0.3$
F1, corresponding to gains of $15.0$ and $13.5$ points over standalone
training in mIoU and F1, respectively. Relative to CIRKD, JEDI provides
additional gains of $1.8$ mIoU and $2.2$ F1 points. For MiT-B2, the
standalone model achieves $56.0\pm0.5$ mIoU and $68.5\pm0.4$ F1, while
CIRKD reaches $68.0\pm0.4$ mIoU and $79.4\pm0.3$ F1. JEDI obtains
$69.5\pm0.4$ mIoU and $81.5\pm0.3$ F1, yielding improvements of
$13.5$ mIoU and $13.0$ F1 points over standalone training and
$1.5$ mIoU and $2.1$ F1 points over CIRKD.

The results therefore show two consistent trends. First, JEDI provides
the best performance for every student capacity and for both evaluation
metrics. Second, the absolute benefit of distillation decreases as
student capacity increases: the mIoU gain of JEDI over standalone
training decreases from $16.0$ points for MiT-B0 to $15.0$ for MiT-B1
and $13.5$ for MiT-B2. This trend is expected as larger students have
greater capacity to learn useful representations directly, but the
persistent improvement over both standalone training and prior
distillation methods indicates that the transferred teacher
representation remains beneficial even when student capacity increases.

JEDI also substantially narrows the gap to the teacher while using
orders of magnitude fewer parameters. The fine-tuned I-JEPA teacher
contains $639.12$M parameters and achieves $70.0\pm0.3$ mIoU and
$84.3\pm0.5$ F1. In comparison, JEDI reaches $68.0\pm0.4$ mIoU with
only $4.04$M parameters for MiT-B0, leaving a $2.0$-point mIoU gap to
the teacher while using approximately $158\times$ fewer parameters.
The MiT-B1 and MiT-B2 students use $14.33$M and $28.0$M parameters,
respectively, yet reach $69.0$ and $69.5$ mIoU, reducing the teacher
gap to $1.0$ and $0.5$ points. Their corresponding F1 gaps are also
reduced from $4.3$ points for MiT-B0 to $3.3$ and $2.8$ points for
MiT-B1 and MiT-B2. Thus, increasing student capacity improves the
absolute agreement with the teacher, while JEDI preserves a substantial
accuracy advantage over training the same students without
distillation.

\subsection{Ablation Studies}

\begin{figure}[t]
    \centering
    \includegraphics[width=\columnwidth]{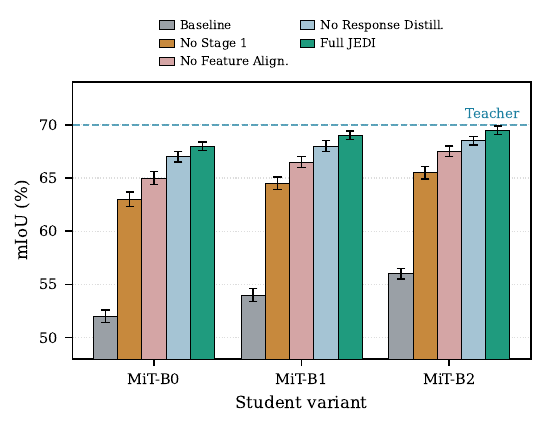}
    \caption{We isolate the contribution of each framework component through
controlled ablations, summarized in Fig.~\ref{fig:ablation} and
Table~\ref{tab:ablation}. The \emph{No Stage 1} configuration removes
the representation-alignment pretraining stage, \emph{No Feature Align.}
sets $\gamma=0$ and therefore removes persistent feature alignment during
Stage~2 and \emph{No Response Distill.} sets $\beta=0$ and removes the
response-distillation term. All ablations are evaluated using the same
five-seed protocol as the main experiments.}
    \label{fig:ablation}
\end{figure}

We isolate the contribution of each framework component through
controlled ablations, summarized in Fig.~\ref{fig:ablation} and
Table~\ref{tab:ablation}. All ablations are run across the same five
seeds as the main results.

\textbf{Stage 1 (representation alignment).}
Removing Stage~1 (\emph{No Stage 1}) and training the student directly
with the Stage~2 objective reduces mIoU at every evaluated capacity.
This indicates that representation alignment provides a useful
initialization rather than functioning only as an optimization
warm-up.

\textbf{Persistent feature alignment ($\gamma=0$).}
The \emph{No Feature Align.} configuration removes the feature-alignment
term during Stage~2 while retaining the Stage~1 initialization. Its
performance is consistently below the full JEDI model, showing that
the representation learned during Stage~1 is not sufficient by itself:
continuing to constrain the student during task adaptation provides an
additional benefit.

\textbf{Response distillation ($\beta=0$).}
The \emph{No Response Distill.} configuration removes the
temperature-scaled response-distillation term while retaining both
representation alignment and persistent feature alignment. The resulting
degradation is smaller than that of the other ablations, but the
consistent improvement of the full JEDI model indicates that teacher
predictions provide complementary information beyond representation
alignment.

\begin{table}[t]
\centering
\caption{Ablation mIoU (\%) on MiT-B0, mean\,$\pm$\,std over five seeds.
Each row removes one component from the full framework.}
\label{tab:ablation}
\setlength{\tabcolsep}{5pt}
\renewcommand{\arraystretch}{1.2}
\begin{tabular}{@{}lc@{}}
\toprule
Configuration & mIoU (\%) \\
\midrule
Standalone (no distillation)         & 52.0\,$\pm$\,0.6 \\
$-$\,Stage 1 (no representation align)& 63.0\,$\pm$\,0.7 \\
$\gamma = 0$ (no persistent feature align) & 65.0\,$\pm$\,0.6 \\
$\beta = 0$ (no response distillation)     & 67.0\,$\pm$\,0.5 \\
\textbf{Full JEDI}                   & \textbf{68.0\,$\pm$\,0.4} \\
\bottomrule
\end{tabular}
\end{table}

\subsection{Choice of Teacher}

To evaluate the effect of teacher choice, we replace I-JEPA with two
alternative ViT teachers: a self-distilled DINO model and a supervised
ViT. Both alternative teachers are fine-tuned for cropland
segmentation using the same protocol as the I-JEPA teacher
(Section~\ref{subsec:models}), before being frozen and distilled into
the student. We then run the same JEDI distillation pipeline for the
MiT-B0 student. Table~\ref{tab:teacher} reports the resulting student
performance. I-JEPA produces the strongest student among the evaluated
teachers, indicating that the teacher representation has a substantial
effect on the effectiveness of the subsequent distillation process.
This experiment isolates teacher choice within the evaluated
setting, since the student architecture, alignment procedure and training
protocol are held fixed.

\begin{table}[t]
\centering
\caption{Effect of teacher choice on the distilled MiT-B0 student
(mIoU \%, mean\,$\pm$\,std over five seeds). The full JEDI pipeline is
run identically for each teacher.}
\label{tab:teacher}
\setlength{\tabcolsep}{5pt}
\renewcommand{\arraystretch}{1.2}
\begin{tabular}{@{}lc@{}}
\toprule
Teacher & Student mIoU (\%) \\
\midrule
Supervised ViT   & 54.5\,$\pm$\,0.6 \\
DINO             & 36.0\,$\pm$\,0.5 \\
\textbf{I-JEPA}  & \textbf{68.0\,$\pm$\,0.4} \\
\bottomrule
\end{tabular}
\end{table}

\subsection{Efficiency Analysis}

Table~\ref{tab:efficiency} characterizes
accuracy relative to computational cost. We report parameter count,
GMACs and NVIDIA A6000 latency, using parameter count and GMACs as
hardware-agnostic indicators of computational footprint. The latency
measurement provides a reference on a high-end GPU rather than a direct
measurement of embedded-device performance. Relative to the teacher, JEDI-B0 reduces
the parameter count by more than two orders of magnitude and reduces
GMACs from $557.5$ to $5.8$ while retaining $68.0$ mIoU. These reductions
substantially improve the computational feasibility of the model under
resource constraints, although actual deployment performance will depend
on the target hardware and runtime.

\input{effc_table}

\subsection{Qualitative Results}
\label{subsec:qualitative}

\begin{figure*}[t]
    \centering
    \includegraphics[width=\textwidth]{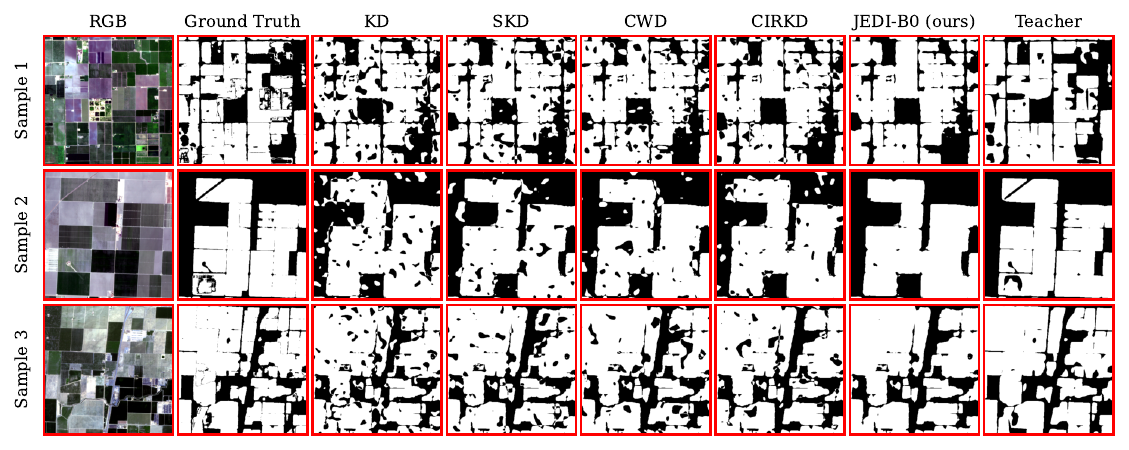}
    \caption{Qualitative comparison on CalCROP21. Columns: RGB input,
    ground truth, four prior distillation baselines (KD, SKD, CWD,
    CIRKD), JEDI-B0 and the fine-tuned teacher. Rows show three
    representative test samples. JEDI-B0 recovers field-level structure
    closely matching the teacher, while prior methods leave larger
    contiguous regions misclassified, consistent with their lower mIoU
    in Table~\ref{tab:main}.}
    \label{fig:qual}
\end{figure*}

Fig.~\ref{fig:qual} shows representative predictions on three test
tiles, selected at random from the test set to avoid
cherry-picking, comparing ground truth against KD, SKD, CWD, CIRKD,
JEDI-B0 and the teacher. The prior distillation methods recover the coarse field
layout but leave larger contiguous regions misclassified, particularly
along field boundaries and within spectrally ambiguous interior patches.
JEDI-B0 produces predictions that more closely track the teacher and
recover coherent field geometry despite its substantially smaller
parameter count. This qualitative behavior is consistent with the
quantitative ordering in Table~\ref{tab:main}.

\section{Discussion and Limitations}
\label{sec:discussion}
JEDI helps most when student capacity is limited. MiT-B0 gains
$16.0$ points over standalone training, while B1 and B2 gain less.
This makes sense: with limited capacity, task supervision alone often
cannot recover the semantic structure already present in the teacher. The ablations show the two stages play different roles. Stage~1
gives a representation-level initialization. Without continued
alignment in Stage~2, however, this initialization erodes as the
student specializes to segmentation. Response distillation adds a
smaller extra gain, showing the teacher's output distribution carries
information beyond the aligned representation. Together, these
results support combining both supervision types under aggressive
compression. \\Three limitations bound our findings. First, we address only binary
cropland segmentation; whether persistent alignment transfers to
multi-class crop mapping remains open. Second, all experiments use one
dataset and region, CalCROP21 over California's Central Valley, so
broader geographic or sensor generalization is untested. Finally,
NDVI-guided selection reduces the temporal stack to a single
acquisition, discarding trajectory information that could help
distinguish spectrally similar crops.\\ \\ \\ 

\section{Conclusion}We propose JEDI, a cross-architecture distillation framework for
transferring a large I-JEPA representation to a compact segmentation
model. The key design is to align the student's terminal-stage
representation with the teacher's token space before task adaptation and
to retain this alignment objective during subsequent segmentation
training. This persistent constraint complements supervised segmentation
and response distillation by reducing the drift of the student's
representation as it specializes to the task. Experiments on CalCROP21 dataset show that the benefit is particularly pronounced
under aggressive compression. The $4.04$M-parameter JEDI-B0 reaches
$68.0$ mIoU, compared with $52.0$ mIoU for the standalone student and
$70.0$ mIoU for the approximately $639$M-parameter teacher. The ablations
show that both representation alignment before task adaptation and its
persistence during adaptation contribute to the final performance, with
persistent feature alignment producing the largest degradation when
removed. Across student capacities, JEDI also provides a favorable
accuracy--compute trade-off, substantially reducing parameter count and
GMACs while retaining most of the teacher's segmentation performance.

\noindent \small {\textbf{Acknowledgements.}} The work was partially supported by the Agriculture and Food Research Initiative Competitive Grant no. 2020-69012-31914 from the USDA National Institute of Food and Agriculture, and by the National Science Foundation under grant no. 2431569, grant 2524228, and the CREST Center for Multidisciplinary Research Excellence in CyberPhysical Infrastructure Systems (MECIS) grant no. 2112650. The views and conclusions in this paper are those of the authors and should not be interpreted as representing any funding agencies.

\bibliographystyle{IEEEtran}
\bibliography{references}

\end{document}

%% file: table_main.tex
\begin{table}[t]
\centering
\caption{Main results on CalCROP21. mIoU and F1 are mean\,$\pm$\,std over
five seeds. For each student capacity we compare the non-distilled
baseline, four prior distillation methods adapted to our
teacher--student pair and JEDI. Best result per capacity in
\textbf{bold}.}
\label{tab:main}
\setlength{\tabcolsep}{4pt}
\renewcommand{\arraystretch}{1.2}
\begin{tabular}{@{}llccc@{}}
\toprule
Model & Method & Params (M) & mIoU (\%) & F1 (\%) \\
\midrule
Teacher (I-JEPA) & fine-tuned & 639.12 & 70.0\,$\pm$\,0.3 & 84.3\,$\pm$\, 0.5 \\
\midrule
\multirow{6}{*}{MiT-B0}
  & Standalone            & \multirow{6}{*}{4.04}
      & 52.0\,$\pm$\,0.6 & 66.0\,$\pm$\,0.5 \\
  & KD~\cite{hinton2015distilling}
      &   & 60.5\,$\pm$\,0.5 & 73.0\,$\pm$\,0.4 \\
  & SKD~\cite{liu2019structured}
      &   & 62.8\,$\pm$\,0.5 & 75.2\,$\pm$\,0.4 \\
  & CWD~\cite{shu2021channel}
      &   & 64.7\,$\pm$\,0.5 & 76.8\,$\pm$\,0.4 \\
  & CIRKD~\cite{yang2022cross}
      &   & 65.9\,$\pm$\,0.5 & 77.9\,$\pm$\,0.4 \\
  & \textbf{JEDI (ours)} &
      & \textbf{68.0\,$\pm$\,0.4} & \textbf{80.0\,$\pm$\,0.3} \\
\addlinespace[3pt]
\multirow{6}{*}{MiT-B1}
  & Standalone            & \multirow{6}{*}{14.33}
      & 54.0\,$\pm$\,0.6 & 67.5\,$\pm$\,0.5 \\
  & KD~\cite{hinton2015distilling}
      &   & 62.0\,$\pm$\,0.5 & 74.3\,$\pm$\,0.4 \\
  & SKD~\cite{liu2019structured}
      &   & 64.1\,$\pm$\,0.5 & 76.1\,$\pm$\,0.4 \\
  & CWD~\cite{shu2021channel}
      &   & 66.0\,$\pm$\,0.5 & 77.9\,$\pm$\,0.4 \\
  & CIRKD~\cite{yang2022cross}
      &   & 67.2\,$\pm$\,0.4 & 78.8\,$\pm$\,0.4 \\
  & \textbf{JEDI (ours)} &
      & \textbf{69.0\,$\pm$\,0.4} & \textbf{81.0\,$\pm$\,0.3} \\
\addlinespace[3pt]
\multirow{6}{*}{MiT-B2}
  & Standalone            & \multirow{6}{*}{28.0}
      & 56.0\,$\pm$\,0.5 & 68.5\,$\pm$\,0.4 \\
  & KD~\cite{hinton2015distilling}
      &   & 63.5\,$\pm$\,0.5 & 75.4\,$\pm$\,0.4 \\
  & SKD~\cite{liu2019structured}
      &   & 65.3\,$\pm$\,0.4 & 77.0\,$\pm$\,0.4 \\
  & CWD~\cite{shu2021channel}
      &   & 67.1\,$\pm$\,0.4 & 78.6\,$\pm$\,0.3 \\
  & CIRKD~\cite{yang2022cross}
      &   & 68.0\,$\pm$\,0.4 & 79.4\,$\pm$\,0.3 \\
  & \textbf{JEDI (ours)} &
      & \textbf{69.5\,$\pm$\,0.4} & \textbf{81.5\,$\pm$\,0.3} \\
\bottomrule
\end{tabular}
\end{table}

%% file: effc_table.tex
\begin{table}[t]
\centering
\caption{Efficiency and accuracy. Latency is mean over 100 runs per
$448\times448$ tile at batch~1, FP32, on one RTX~A6000. GMACs are
measured with fvcore and are approximate---attention, GELU and softmax
ops are not counted, so all figures are lower bounds. mIoU is mean over
five seeds.}
\label{tab:efficiency}
\setlength{\tabcolsep}{4pt}
\renewcommand{\arraystretch}{1.2}
\begin{tabular}{@{}lcccc@{}}
\toprule
Model & Params (M) & GMACs$^{\dagger}$ & Lat. (ms) & mIoU (\%) \\
\midrule
Teacher (I-JEPA) & 639.12 & 557.5 & 83.5 & \textit{70.0} \\
JEDI-B2          & 28.00  & 45.9  & 17.0 & \textit{69.5} \\
JEDI-B1          & 14.33  & 11.4  & 7.1  & \textit{69.0} \\
JEDI-B0          & 4.04   & 5.8   & 7.1  & \textit{68.0} \\
\bottomrule
\end{tabular}
\end{table}